\documentclass[journal,twoside,web]{ieeecolor}
\usepackage{generic}
\usepackage{cite}
\usepackage{amsmath,amssymb,amsfonts}
\usepackage{algorithmic}
\usepackage{graphicx}
\usepackage{algorithm,algorithmic}
\usepackage{textcomp}
\usepackage{threeparttable}
\usepackage[table,dvipsnames]{xcolor}
\usepackage{multirow}
\usepackage{tabularx}
\usepackage{booktabs}
\usepackage{hyperref}
\usepackage{mathrsfs}
\usepackage[dvipsnames]{xcolor}
\hypersetup{colorlinks=true, citecolor=blue, anchorcolor=blue}
\def\BibTeX{{\rm B\kern-.05em{\sc i\kern-.025em b}\kern-.08em
    T\kern-.1667em\lower.7ex\hbox{E}\kern-.125emX}}
\begin{document}
\title{DiffSight-Former: Modeling Structural Differences and Temporal Dynamics for Glaucoma Progression Prediction}
\author{Yi Huang, Lei Bi, and Jinman Kim \IEEEmembership{Member, IEEE}
\thanks{Yi Huang and Jinman Kim are with the School of Computer Science, The University of Sydney. (e-mail: yhua0932@uni.sydney.edu.au; jinman.kim@sydney.edu.au)}
\thanks{Lei Bi is with the Institute of Translational Medicine, Shanghai Jiao Tong University. (e-mail: lei.bi@sjtu.edu.cn)}
}

\maketitle

\begin{abstract}
Glaucoma is a leading cause of irreversible blindness worldwide, and early detection from fundus images is critical for effective disease management. While deep learning has achieved promising performance in fundus image analysis, most existing methods rely on single time-point images and therefore fail to capture longitudinal structural and vascular changes that are informative for disease progression. Sequential fundus images collected during clinical follow-up provide valuable temporal information; however, current sequential models often struggle to detect subtle early progression signals and commonly rely on fixed-length inputs or diagnostic cues from already glaucoma images, limiting their clinical utility for early forecasting. 
To address these limitations, we propose DiffSight-Former, a framework for glaucoma progression prediction from sequential fundus images. It incorporates a time-variant feature extraction module based on a fundus-specific foundation model to obtain robust anatomical representations. A multi-structure difference modeling module is introduced to quantify progression-related structural changes in the optic disc/cup region and retinal vasculature. These representations are integrated with temporal interval embeddings and processed by a time-aware Transformer to model disease progression dynamics and estimate the probability of future glaucoma onset.
Experiments were conducted on two longitudinal datasets, SIGF (405 sequences) and GRAPE (263 sequences). On SIGF, DiffSight-Former achieved an AUC of 91.54\% and a sensitivity of 92.16\% for progression prediction. On GRAPE, the proposed method achieved an average accuracy of 87.48\% across three clinical visual-field progression criteria. Compared with existing approaches, DiffSight-Former demonstrates consistently strong performance and improved robustness across different temporal settings, highlighting its potential for longitudinal glaucoma monitoring and early risk prediction.
\end{abstract}


\section{Introduction}
\label{sec:introduction}
\IEEEPARstart{G}laucoma is one of the leading causes of irreversible blindness worldwide \cite{danesh2015}, affecting more than 70 million individuals \cite{blindness2017, projections2014}. Early identification of disease progression is essential for preventing vision loss and guiding timely clinical intervention. Color fundus photography is widely used for glaucoma screening and longitudinal monitoring due to its non-invasive nature and ability to capture structural changes in the optic nerve head and retinal vasculature \cite{fundusimage2022, trends2022}. With the increasing availability of longitudinal imaging data in clinical practice, automated analysis of sequential fundus images has become an important research direction for predicting glaucoma progression \cite{OHTS, AREDS, DeepGF2020miccai, GRAPE, qi2024myopia}. Despite using of longitudinal fundus imaging in clinical practice, accurately predicting glaucoma progression from sequential images remains a challenging problem.

Deep learning techniques have shown promising performance in fundus image analysis, particularly for glaucoma detection and segmentation tasks \cite{Li2020CANet, He2021TMI, Pammi2024BSPC, RadialMRI2025}. However, most existing approaches analyze fundus images at a single time point and therefore cannot capture disease progression patterns that unfold over time \cite{fu2018tmi, Vutukuru2023BSPC, Dipankar2024, G-Risk2023,RDD-Net2024}. In clinical practice, glaucoma progression is typically characterized by gradual structural changes in the optic disc, cup morphology, and retinal vasculature across multiple visits, rather than abrupt diagnostic appearance differences. Sequential imaging data thus provide important temporal cues that may improve progression prediction. Consequently, models designed for static diagnosis may not effectively capture progression-related patterns in longitudinal data.

\begin{figure*}[t]
\centering
\includegraphics[width=15.5cm]{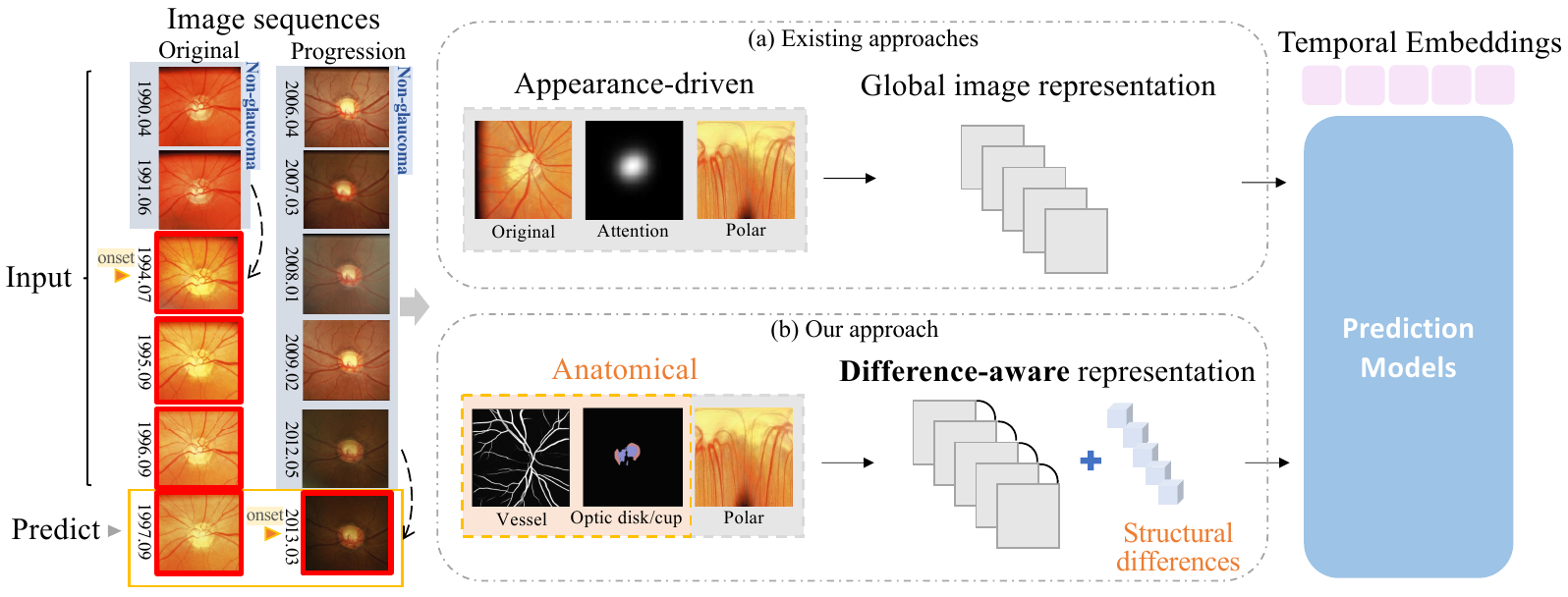}
\caption{
Comparison between existing approaches and the proposed method for glaucoma progression prediction from sequential fundus images. Existing methods primarily rely on global image representations derived from appearance-driven features, which are subsequently aggregated with temporal embeddings for prediction. 
The proposed framework explicitly models inter-visit structural differences by leveraging anatomical priors, including optic disc/cup and retinal vasculature, to capture progression-related changes. 
These difference-aware representations, combined with temporal interval information, enable more effective prediction of early glaucoma progression.
} \label{fig1-1}
\end{figure*}

Recent studies have attempted to model longitudinal fundus sequences using recurrent neural networks or Transformer-based architectures \cite{2017jose, saha2023,2021Song, pmpod2025}. Despite these advances, two important challenges remain. First, subtle structural changes associated with early glaucoma progression are often difficult to capture using global image representations. Second, many sequential models rely on fixed-length inputs or include already glaucomatous images in the input sequences \cite{DeepGF2020miccai, 2022DRMA, Glim2023tmi, MSTFormer2025jbhi}, which may bias predictions toward diagnostic appearance cues rather than true progression dynamics. These approaches typically encode each image as a global representation and rely on temporal aggregation to infer disease evolution. However, such strategies may overlook localized structural differences between visits that are critical for identifying early progression.

Recently, foundation models (FMs) pretrained on large-scale retinal image datasets have shown strong capability in learning generalizable representations for ophthalmic image analysis \cite{RETFound2023,VisionFM2024,eyeFM2025,eyeCLIP2025}. These models capture rich anatomical and semantic features that may benefit longitudinal modeling of disease progression. However, effectively leveraging such representations for sequential progression forecasting remains challenging, particularly when subtle structural changes must be distinguished from non-pathological appearance variations across visits. Nevertheless, effectively leveraging foundation model representations for progression forecasting requires mechanisms that can highlight subtle inter-visit structural changes while suppressing appearance variations unrelated to disease progression.

To address these challenges, we propose DiffSight-Former, a framework designed to model structural differences and temporal dynamics for glaucoma progression prediction from sequential fundus images. 
As illustrated in Fig. \ref{fig1-1}, existing approaches primarily rely on global appearance representations aggregated over time, which may overlook subtle progression cues when pathological changes are not visually apparent. 
In contrast, the proposed framework explicitly models inter-visit structural differences in anatomically relevant regions, enabling the extraction of progression-related signals even under clinically early-stage conditions with non-glaucoma inputs. 
This design allows the model to capture disease evolution more effectively while accounting for irregular temporal intervals in longitudinal data.

Specifically, we employ a Time-variant Feature Extraction (TFE) module that leverages a fundus-specific foundation model to extract dynamic representations that capture both semantic and anatomical characteristics of retinal structures across visits. A Multi-Structure Difference Modulation (MDM) module is introduced to detect subtle inter-scans variations by comparing semantic features from multiple retinal structures, thereby enhancing sensitivity in detecting early pathological changes. Finally, both temporal interval information and inter-image difference features are embedded into a Transformer-based architecture to explicitly associate structural changes with temporal dynamics, as such enabling to model irregular follow-up intervals and forecast glaucoma progression more effectively. This design allows the framework to capture progression dynamics under irregular follow-up intervals commonly observed in real-world clinical data.
Based on the above considerations, this work makes the following contributions:
\begin{itemize}

\item
{We propose DiffSight-Former, which explicitly models inter-visit structural changes in sequential fundus images to improve early glaucoma progression prediction.}

\item
{We design multi-structure difference modeling for capturing subtle anatomical changes, integrating information from the optic disc/cup region, retinal vasculature, and fundus appearance to better characterize progression-related structural variations. }

\item
{We embed both temporal interval information and inter-scan difference features into a time-aware Transformer, enabling effective modeling of disease dynamics under irregular clinical follow-up intervals.}

\item
{Experiments on two independent datasets demonstrate that the proposed method achieves strong performance and improved robustness across different temporal settings and clinical progression criteria.}

\end{itemize}


\section{Related Work}
We have categorized the related studies into (1) temporal modeling for medical image sequences; and (2) early glaucoma prediction.

\subsection{Temporal Modeling on Medical Image Sequences}
Research across multiple clinical domains has demonstrated the importance of leveraging sequential data to monitor disease progression and forecast clinical outcomes. For instance, longitudinal modeling has been used to predict Alzheimer’s disease progression from structural magnetic resonance imaging (MRI) \cite{ADP2025NN}, track femoral shape changes associated with hip osteoarthritis using CT \cite{hipct2024miccai}, and forecast knee osteoarthritis progression from X-Ray imaging \cite{oai2025miccai}.
In ophthalmology, sequential imaging has also been applied to, for instance, predict the progression in age-related macular degeneration (AMD), diabetic retinopathy (DR), and myopia \cite{chakravarty2024tmi, dmitri2024tmi, drprogression2024miccai,qi2024myopia}. A variety of sequential architectures have been explored to capture temporal dependencies. Early approaches primarily employed recurrent neural networks (RNNs), including long short-term memory (LSTM) and gated recurrent units (GRUs), which model temporal dependencies by sequentially updating hidden states \cite{convlstm2015, rnn2019tvcg, datagru2020aaai, rnn2024ASC}. These models effectively capture coarse temporal trends but struggle with long-range dependencies, spatial feature preservation, and irregular sampling, which reduce their suitability for complex medical image sequences.

More recently, Transformers have emerged as a powerful architecture for sequence modeling, owing to their ability to capture long-range dependencies via self-attention mechanisms. In medical imaging, Transformer-based models have been applied to diverse tasks such as in image classification, lesion segmentation, detection, reconstruction, synthesis, registration, and clinical report generation \cite{transformers2023survey,transseg2024survey, chowdary2024miccai, recon2024tmi, Wang_2023_CVPR}. For sequential medical data, temporal Transformers have been explored in applications including video analysis, longitudinal radiology studies, and electronic health records (EHRs) \cite{RHRs2023miccai, PATrans2024tmi,timeserial2024miccai, hong2024wacv, dengao2025ci}.

To encode time information directly from the data, several works have extended Transformers with time-aware mechanisms, incorporating explicit temporal embeddings or event-time modeling into the self-attention layer. For example, Wu et al. \cite{PATrans2024tmi} proposed the Pattern-Aware Transformer (PATrans), which introduces a hierarchical tokenization strategy featuring a global-local dual-path pattern-aware cross-attention mechanism to handle video object detection or 3D segmentation across sequential medical images. Li et al. \cite{linips2023} introduced ViTST, an approach for modeling irregularly sampled time series by transforming them into graph images and leveraging pre-trained vision transformers for classification. Xiao et al. \cite{xtsformer2025aaai} proposed XTSFormer, a cross-temporal‑scale Transformer tailored for irregularly clinical events modeling and prediction from EHRs. Li et al. proposed \cite{Uniformer2023} UniFormer, a unified architecture that integrates the strengths of convolution and self-attention within a concise Transformer framework, performing well in a variety of image and video task.
However, these methods are primarily designed for regularly sampled data and do not explicitly model subtle structural changes across irregular clinical visits, limiting their effectiveness for longitudinal fundus progression forecasting.

\subsection{Early Glaucoma Prediction} 
Glaucoma presents unique challenges for disease forecasting because its structural deterioration is typically slow, subtle, and often clinically indistinguishable in early stages, making short-term changes difficult to detect. Earlier studies primarily employed CNNs and LSTM networks to capture spatial features and temporal dynamics. In 2020, Li et al. \cite{DeepGF2020miccai} introduced the SIGF dataset, the first sequential fundus dataset for glaucoma forecasting. Along with this dataset, they proposed DeepGF, which combined CNN-based spatial feature extraction with LSTM-based temporal modeling. They modeled glaucoma progression by taking a sequence of five chronological fundus images as input and predicting the label of the next clinical visit, thereby formulating forecasting as a next-step progression classification problem.

Building on this, Hu et al. \cite{Glim2023tmi} developed GLIM-Net, the first Transformer-based framework applied to irregularly sampled SIGF sequences. By incorporating time positional encoding and time-sensitive multi-head self-attention, GLIM-Net improved forecasting accuracy and enabled prediction conditioned on a specific future time. However, the model represented each fundus image as a single global token, which restricts its ability to model local structural changes, such as early optic cup enlargement or vessel displacement, that are critical predictive cues for identifying subtle progression trends.
Later, Holste et al. \cite{LTSA2024npj} introduced the Longitudinal Transformer for Survival Analysis (LTSA), which modeled disease time-to-event outcomes from irregularly sampled fundus image sequences, outperforming single-image baselines for forecasting slow, progressive eye diseases including primary open-angle glaucoma and AMD on OHTS \cite{OHTS} and AREDS \cite{AREDS} datasets.

Most recently, Yang et al. \cite{MSTFormer2025jbhi} proposed MST-former, which incorporated space-time positional encoding, multi-scale spatial feature extraction, and time-aware temporal attention to better capture intra-image details and inter-visit progression cues. In addition, a temperature-controlled Balanced Softmax loss was introduced to address extreme class imbalance issues.

Despite these advances, several limitations remain. Existing studies typically employ fixed-length input sequences and often include already glaucoma images in the input, which limits their clinical relevance for early forecasting. More importantly, most methods represent each image as a global feature and rely on temporal aggregation to infer progression, making them less sensitive to subtle, localized structural changes that characterize early-stage disease. As a result, predictions are often dominated by the most recent image, leading to biased learning and reduced generalizability.

\begin{figure*}[t]
\centering
\includegraphics[width=17cm]{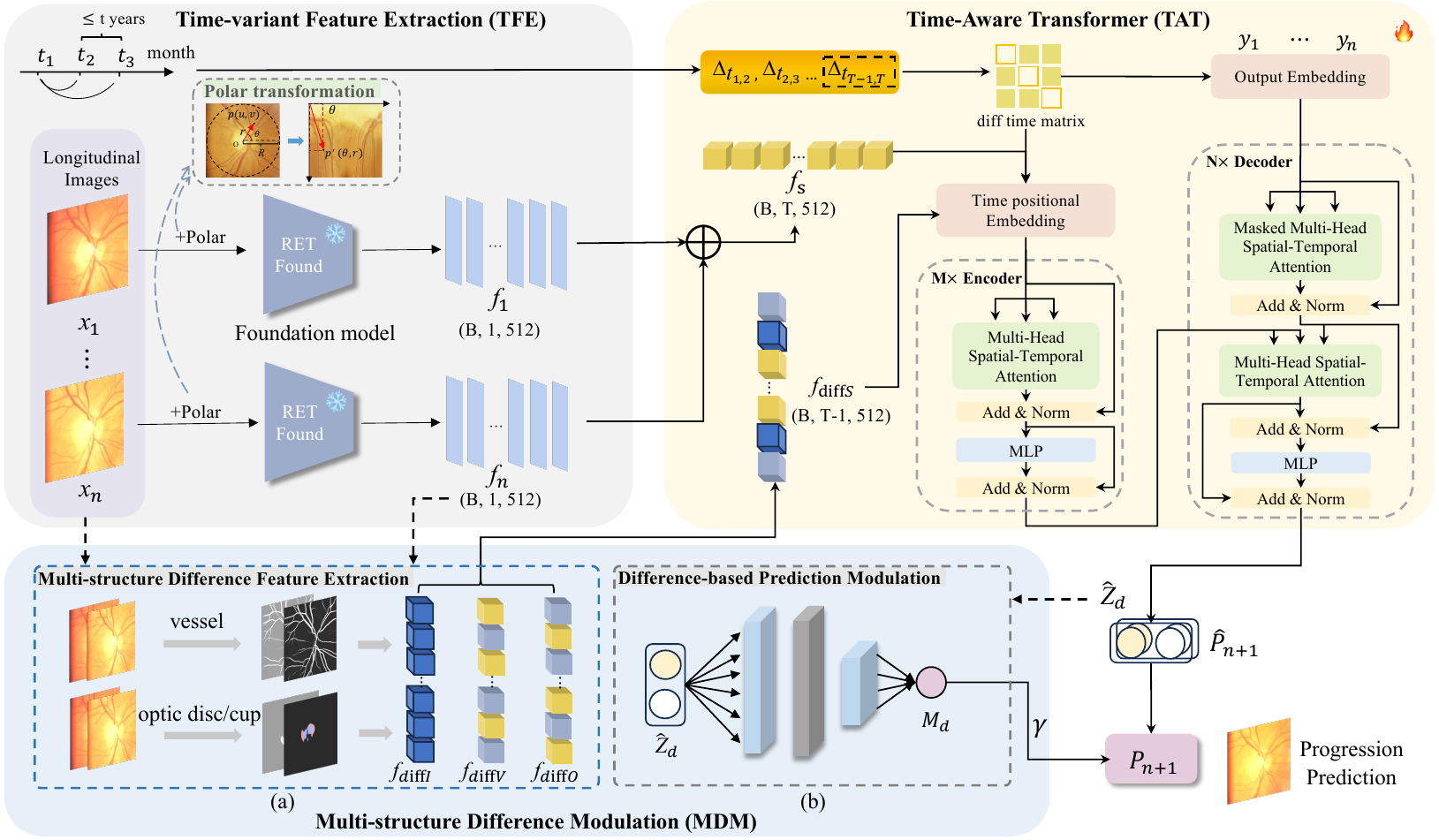}
\caption{{Overview of the proposed DiffSight-Former, which consists of three modules: Time-Variant Feature Extraction (TFE), Multi-structure Difference Modulation (MDM) and Time-Aware Transformer (TAT). In the TFE module, sequential fundus images are first polar-transformed and then encoded using a fundus-specific foundation model (RETFound) to obtain time-variant image features. In MDM module (a), the vessel and optic disc/cup maps are generated, and their corresponding difference features are computed to characterize structural and vascular changes. In the TAT module, the time interval between images is encoded, and three types of inputs are fused: (1) the original image features extracted by TFE, (2) the multi-structure difference features from MDM, and (3) the temporal interval embeddings. These fused representations are processed by a time-aware Transformer to model progression dynamics. Finally, the MDM module (b) modulates the Transformer’s output to adjust the predicted probability of future glaucoma progression, yielding the final prediction.}} \label{fig3-1}
\end{figure*}

\section{Proposed Method}
Our proposed method is shown in Fig. \ref{fig3-1}., which consists of three main components: (A) Time-Variant Feature Extraction (TFE), (B) Multi-structure Difference Modulation (MDM), and (C) Time-Aware Transformer (TAT). Given a sequence of longitudinal fundus images $\{x_1, x_2, \ldots, x_T \space (T<=5)\}$ with the time stamps $\{t_1, t_2, \ldots, t_T\}$ and label $\{y_1, y_2, \ldots, y_T \}$ , TFE extracts time-variant representations using a fundus-specific foundation model. MDM computes multi-structure difference features from original images as well as vessel and optic disc/cup maps. Finally, TAT incorporates temporal interval embeddings and integrates all features through a time-aware Transformer backbone, producing a probability prediction $P_{T+1}$ for future glaucoma onset. 

\subsection{Time-variant Feature Extraction Module}
The first stage aims to obtain robust and semantically rich representations of sequential fundus images. The input images are normalized to a shape of $ B\times3\times 224\times 224$. $ B$ stands for batch size. Following \cite{fu2018tmi}, each input image in the sequence $\{x_1, x_2, \ldots, x_T\}$ is first transformed by polar coordinates, which enlarge the optic disc/cup region and normalize structural variations. The process of polar coordinate transformation is shown in Fig. \ref{fig3-1}. For each pixel $p(u,v)$ in the original image, its position is transformed to $p'(\theta,r)$ as follows:
\begin{equation}
\left\{\begin{aligned}
    r  &=\sqrt{u^{2}+v^{2}} \\
\theta &=\tan ^{-1}\left(\frac{v}{u}\right)
\end{aligned}\right.
\end{equation}
where $u,v$ represent the position in the original coordinate system. $\theta, r$ represent the angle and radius in the polar coordinate system.

Then, the polar images are processed by RETFound \cite{RETFound2023}, a large-scale fundus-specific foundation model pre-trained on ImageNet-1k and 1.6 million retinal images via the self-supervised Mask Autoencoder. With an encoder-decoder architecture, RETFound uses 24 transformer blocks as the encoder with an embedding vector of size 1024, and 8 transformer blocks as the decoder and an embedding vector size of 512. Here, we load its pre-trained weights and freeze all parameters for the time-variant feature extraction.
After all the polar image sequences are extracted, image features $\{f_1, f_2, \ldots, f_T\}$ are obtained with the shape of $ B\times1\times 512$ respectively. They are then concatenated to form the sequence feature $f_s$ in the shape of $ B\times T\times 512$.

\subsection{Multi-structure Difference Modulation Module}
To capture reliable cues of glaucoma progression, we compute difference features across visits at multiple structures. As glaucoma develops, the most salient structural alterations occur in the optic cup/disc, where enlargement of the cup-to-disc ratio is an indicator of disease progression. In addition, vascular patterns in the peripapillary region usually shift due to structural remodeling of the optic nerve head. They provide distinct information at different spatial scales: vessels reflect fine-grained, high-frequency changes, while optic disc/cup morphology reflects coarser, low-frequency alterations. Here the module is composed of 2 sub-modules: (1) Multi-structure Difference Feature Extraction, as illustrated in Fig. \ref{fig3-1} MDM (a), (2) Difference-based Prediction Modulation, as illustrated in Fig. \ref{fig3-1} MDM (b).

\subsubsection{Multi-structure Difference Feature Extraction}
To get the vessel map and optic cup and disk map corresponding to each input image, we utilize AutoMorph \cite{AutoMorph2022} to process the original images. As a fully automated method for quantifying retinal morphology, AutoMorph can be applied to image quality assessment, anatomical tissue segmentation, and clinically relevant index measurement, effectively performing quantitative analysis of retinal morphology under different human health conditions. Using AutoMorph, the sequence images $\{x_1, x_2, \ldots, x_T\}$ are processed to get the vessel maps $\{v_1, v_2, \ldots, v_T\}$ and the optic cup and disk maps $\{o_1, o_2, \ldots, o_T\}$ respectively. Feature embeddings are extracted from these maps, then compared across visits to compute temporal difference features together with the polar images features $\{f_1, f_2, \ldots, f_T\}$. These maps are in the same way firstly processed by RETFound to get the vessel features $\{fv_1, fv_2, \ldots, fv_T\}$ and the optic cup and disk features $\{fo_1, fo_2, \ldots, fo_T\}$. After this, the difference features are calculated as follows:
\begin{equation}
\left\{\begin{aligned}
\Delta f_{{t}}   &=f_{t}-f_{t-1} \\
\Delta f_{v_{t}} &=fv_{t}-fv_{t-1} \\
\Delta f_{o_{t}} &=fo_{t}-fo_{t-1} \\
\end{aligned}\right.
\end{equation}
in which $ t= 2,3,...,T$. So for a sequence with $T$ images, there are $T-1$ vessel and optic cup and disk difference features. The overall polar image difference features $f_{diffI}$, vessel difference features $f_{diffV}$, and optic cup and disk difference features $f_{diffO}$ are combined by the following equations:
\begin{equation}
f_{diffI}= \sum_{t=2}^{T} \Delta f_{{t}},\space
f_{diffV}= \sum_{t=2}^{T} \Delta f_{v_{t}},\space
f_{diffO}= \sum_{t=2}^{T} \Delta f_{o_{t}}
\end{equation}
By explicitly modeling these changes in the original fundus images, the network can capture subtle but clinically meaningful differences that may be overlooked when only global embeddings are used.


\subsubsection{Difference-based Prediction Modulation} 
Since the difference features contain change information, they are used to refine the probability prediction output from the TAT module. 

Specifically, a scalar modulation factor derived from differences features is applied to adjust the raw prediction. Denote $\mathbf{L} \in \mathbb{R}^{B \times T+1 \times C}(C=2)$ as the class-logits emitted by the TAT module, $Z_d$ is the last token encodes the difference vector.
\begin{equation}
\begin{array}{cc}
\mathbf{Z}_{d}=\mathbf{L}_{:, T,:} \in \mathbb{R}^{B \times C}
\end{array}
\end{equation}
A lightweight MLP predicts a scalar modulation factor,

\begin{equation}
\begin{aligned}
\operatorname{MLP}_{\text {reg }}(x) & =W_{2} \sigma\left(W_{1} x+b_{1}\right)+b_{2}  \\
 m & =\operatorname{MLP}_{\text {reg }}\left(\mathbf{z}_{d}\right) \in \mathbb{R}^{B} 
\end{aligned}
\end{equation}

with $W_{1} \in \mathbb{R}^{32 \times C}, W_{2} \in \mathbb{R}^{1 \times 32}$ and GELU activation $\sigma$.
A learnable scalar $\gamma$, initialised to 0.5, broadcasts $m$ over the full logit tensor:
\begin{equation}
\mathbf{L}^{\left.\right|^{*}}=\mathbf{L}+\gamma m \otimes \mathbf{1}_{3 \times C}, \quad m \otimes \mathbf{1}_{3 \times C} \in \mathbb{R}^{B \times 3 \times C} \\
\end{equation}
Thus logits of each sequence are adaptively boosted or attenuated in proportion to the learned difference magnitude. When greater structural changes are detected, the probability of future glaucoma is increased accordingly. 
Through this two-stage design, MDM explicitly aligns image-derived structural progression cues with model outputs, ensuring that predictions are not only data-driven but also grounded in clinical markers of glaucomatous change.


\subsection{Time-Aware Transformer}
In this module, based on the Transformer \cite{Attention2017} structure, we embedded the time gap and the difference features with the fundus features extracted by the foundation model to output the prediction probability. 

\subsubsection{Time positional embedding} 
Since the Transformer lacks sequence information, we need to manually inject position into the input image feature vector. However, the original position encoding assumes no temporal differences between images. Our sequential images are acquired at different times, so temporal information is incorporated into the position encoding. To explicitly incorporate irregular follow-up intervals into the Transformer, we extend the sinusoidal positional encoding scheme \cite{Attention2017} to encode time intervals between visits. Suppose each images from the sequence $\{x_1, x_2, \ldots, x_T\}$ is taken at a corresponding time point $\{t_1, t_2, \ldots, t_T\}$, measured in months. For each visit $i$, we compute its relative time interval with respect to the last visit:
\begin{equation}
    \Delta t_{i} = t_{T}-t_{i}, \space i = 1,2, \dots T-1
\end{equation}
Denote the embedding dimension is $E$. For each time interval $\Delta t_{i}$, the encoding vector is defined as:
\begin{equation}
\begin{aligned}
TPE(\Delta t_{i},2n) &= sin(\frac{\Delta t_{i}}{10000^{2n/E}})\\
TPE(\Delta t_{i},2n+1) &= cos(\frac{\Delta t_{i}}{10000^{2n/E}})
\end{aligned}
\end{equation}
where $n = 0,1,2,\dots, E/2-1$. Thus, for each sequence we obtain a matrix:
\begin{equation}
TPE = [TPE(\Delta t_1),TPE(\Delta t_2),\dots, TPE(\Delta t_T)]\in \mathbb{R}^{T\times E}
\end{equation}
In practice, we precompute a positional encoding matrix $P\in \mathbb{R}^{T\times E}$, where $L$ is the maximum sequence length. Each row $P[\mathscr{l}]$ corresponds to the encoding of a time interval $\mathscr{l}$. Given a batch of visit intervals ${\Delta t}$, we simply gather the corresponding rows of $P$, producing the final time encoding tensor:
\begin{equation}
Z_{time} = P[\Delta t_1, \Delta t_2,\dots, \Delta t_T]\in \mathbb{R}^{B \times T\times E}
\end{equation}
where $B$ is the batch size. These encodings are added to the fundus feature embeddings before feeding into the Transformer encoder, allowing the model to differentiate visit intervals of varying lengths and capture time-dependent progression patterns.

\subsubsection{Feature difference embedding}
In addition to temporal embeddings, we incorporate feature difference embeddings to explicitly model inter-visit structural changes, which are often more informative for glaucoma progression than absolute image features. The sequential fundus embeddings extracted by TFE module is $f_{s}$. The difference features of different structure are $f_{diff_{I}}$, $f_{diff_{V}}$, and $f_{diff_{O}}$. To reflect their varying contributions, we introduce learnable weights $w_{I},w_{V},w_{O} \in \mathbb{R}$, normalized by a softmax constraint:
\begin{equation}
    w_{I}+w_{V}+w_{O} = 1
\end{equation}
The final multi-structure difference feature is then computed as:
\begin{equation}
    f_{diffS} = w_{I}\cdot f_{diff_{I}} +w_{V}\cdot f_{diff_{V}}+w_{O}\cdot f_{diff_{O}}
\end{equation}
Finally, the multi-structure difference feature $f_{diffS}$ is concatenated with the time-aware fundus embeddings $F^{'} \in \mathbb{R}^{B \times T\times E}$:
\begin{equation}
   X = Contat(F^{'},f_{diff}) \in \mathbb{R}^{B \times T+1 \times E}
\end{equation}
This enriched representation $X$ preserves both spatial features and multi-structure relative progression cues, serving as the input to the Transformer encoder.

\subsubsection{Encoder-decoder structure with Time-aware attention}
Following the vanilla Transformer structure, there are $M$ encoders and $N$ decoders stacked together. Each encoder consists of a time-aware multi-head self-attention layer and a feed-forward network, accompanied by a residual connection and layer normalization. Each decoder has three sub-layers: a masked time-aware multi-head self-attention layer, a time-aware multi-head self-attention layer and a feed-forward network, also with a residual connection and layer normalization. 

Specifically, we extend the standard multi-head self-attention mechanism by incorporating a time-related weighting matrix that encodes the temporal interval between visits. This ensures that attention weights reflect not only feature similarity but also the relative distance in time between observations. Given input features $X \in \mathbb{R}^{B \times (T+1)\times d_{model}}$ for a batch of size $B$, sequence length $T$, and feature dimension $ d_{model}$, we compute queries, keys, and values via linear projections:  
\begin{equation}
    Q=XW_Q,\space K=XW_K,\space V=XW_V
\end{equation}
where $W_Q,W_K,W_V \in \mathbb{R}^{d_{model} \times d_{model}}$. We then split them into $h$ heads with per-head dimension  $ d_{k}= d_{model}/h$:
\begin{equation}
    Q=[Q^{(1)},Q^{(2)},\dots,Q^{(h)}],\space Q^{(u)} \in \mathbb{R}^{B \times (T+1)\times d_{k}}
\end{equation}
and similarly for $K,V$. To account for temporal irregularity, we introduce a time weighting matrix $\Delta \in \mathbb{R}^{N \times (T+1)\times (T+1)}$, where $\Delta_{ij}$ encodes the time interval between visit $i$ and $j$. A learnable function $\Phi(\cdot)$ transforms this into a multiplicative bias $T=\Phi(\Delta)$.
\begin{equation}
\widetilde{S}^{(u)}=\left(Q^{(u)}(K^{(u)})^{\top}\right)\odot \Phi({\Delta})
\end{equation}

For each head $u$, the normalized attention scores are given by the dot product between queries and keys:
\begin{equation}
   H^{(u)}=Softmax(\frac{\widetilde{S}^{(u)}}{\sqrt{d_{k}}})V^{(u)} \in \mathbb{R}^{N \times T \times T}
\end{equation}
All heads are concatenated and projected:
\begin{equation}
   MHATT(X)=[H^{(1)}|| H^{(2)}||\dots||H^{(h)}]
\end{equation}
Finally, the output of the time-aware multi-head self-attention layer is:
\begin{equation}
   Y=LayerNorm(X+MHATT(X))
\end{equation}
The final output of the time-aware transformer is $\mathbf{Z}_{d}||P_{T+1}$.
\subsection{Loss function}
In our task, the progression outcome is binary, so we adopt the cross-entropy loss to optimize the glaucoma progression prediction model, which can be formulated as:
\begin{equation}
    \mathcal{L}_{C E}=-\frac{1}{B} \sum_{i=1}^{B}\left[y_{i} \log \left(\hat{p}_{i}\right)+\left(1-y_{i}\right) \log \left(1-\hat{p}_{i}\right)\right]
\end{equation}
where $y_{i} \in \{{0,1\}}$ denotes the ground truth label for the 
$i$-th sequence, $ \hat{p}_{i}$ denotes the predicted probability of the $i$-th sequence.

\section{Experiments}
\subsection{Datasets}
We evaluate the performance of DiffSight-Former compared to the state-of-the-art counterpart using two sequential glaucoma datasets.

\begin{figure}[t]
\centering
\includegraphics[width=9cm]{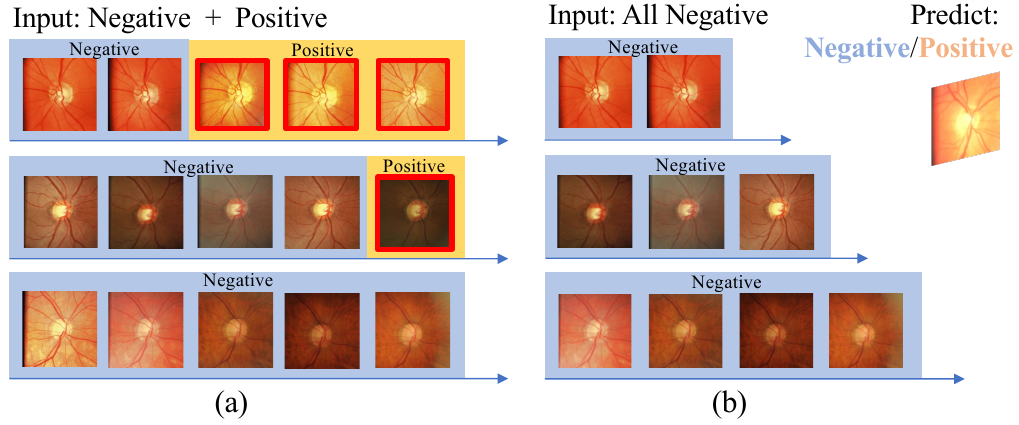}
\caption{Different input sequences. (a) Input sequences in previous studies: the negative and positive glaucoma images are all used to predict with fixed input length. (b) Input sequences in our method: only negative glaucoma images are used to predict, with flexible input length. } \label{fig4-4}
\end{figure}

\begin{table}[ht]
\vspace{-3mm}
\caption{{Data distribution of the sequences from the SIGF dataset.}}\label{tab1}
\setlength{\tabcolsep}{3mm}
\centering
\begin{tabular}{lccc}
\hline
        Label                 & Train & Validation & Test \\ \hline
\multicolumn{4}{c}{T=2}            \\ \hline
Progression              & 221   & 26         & 62   \\
Non-progression          & 221   & 26         & 62    \\ 
Total                    & 442   & 52         & 124   \\ \hline
\multicolumn{4}{c}{T=3}            \\ \hline
Progression              & 603   & 68         & 163   \\
Non-progression          & 603   & 68         & 163    \\ 
Total                    & 1206  & 136        & 326   \\ \hline
\multicolumn{4}{c}{T=4}            \\ \hline
Progression              & 1612   & 180         & 251   \\
Non-progression          & 1612   & 180         & 251    \\ 
Total                    & 3224   & 360         & 502   \\ \hline
\end{tabular}
\vspace{-2mm}
\end{table}

\begin{table}[]
\vspace{-3mm}
\caption{{Data distribution of the sequences from the GRAPE dataset.}}\label{tab2}
\setlength{\tabcolsep}{2.5mm}
\centering
\begin{tabular}{clccc}
\hline
Type                  & \multicolumn{1}{c}{Label} & Eyes Number & Training & Test \\ \hline
\multirow{3}{*}{PLR2} & Progression               & 26                                                 & 20       & 6    \\
                      & Non-progression           & 170                                                & 136      & 34   \\
                      & Total                     & 196                                                & 156      & 40   \\ \hline
\multirow{3}{*}{PLR3} & Progression               & 8                                                  & 6        & 2    \\
                      & Non-progression           & 188                                                & 151      & 37   \\
                      & Total                     & 196                                                & 156      & 40   \\ \hline
\multirow{3}{*}{MD}   & Progression               & 16                                                 & 12       & 4    \\
                      & Non-progression           & 180                                                & 144      & 36   \\
                      & Total                     & 196                                                & 156      & 49   \\ 
\hline 
\end{tabular}
\end{table}

\begin{table*}
\setlength{\tabcolsep}{3.0mm}
\renewcommand{\arraystretch}{1.1}
\centering
\caption{{Comparison of the proposed method with other methods on SIGF dataset.(unit:\%)}}\label{tab3}
\begin{tabular}{cccccccccc}
\toprule
\multirow{3}{*}{Method} & \multirow{3}{*}{Year} & ACC  & SEN & SPE & AUC & ACC & SEN & SPE & AUC \\ 
                                                 \cmidrule(r){3-6}        \cmidrule(r){7-10} 
                        &                       & \multicolumn{4}{c}{Progression ($T$=2)} & \multicolumn{4}{c}{Augmentation ($T$=5)} \\
\midrule
ConvLSTM \cite{convlstm2015} & NeurIPS2015 & 57.26 & \underline{67.74} & 46.77 & \multicolumn{1}{c|}{55.25} & 60.23 & 60.78 & 55.83 & 67.15 \\
DeepGF \cite{DeepGF2020miccai} & MICCAI2020 & 60.48 & 66.13 & 58.06 & \multicolumn{1}{c|}{65.99}  & 66.90 & 66.67 & 66.90 & 71.28 \\
Uniformer\cite{Uniformer2023} & TPAMI2023 & \underline{64.52} & 64.52 & \underline{64.52} & \multicolumn{1}{c|}{67.04} & 79.03 &68.80 & 78.34 & 74.44\\
GLIM-Net \cite{Glim2023tmi}& TMI2023 & 60.83 & 61.29 & 60.34 & \multicolumn{1}{c|}{72.16}  & 80.29 & 80.39 & 80.28 & 87.38 \\
MST-Former \cite{MSTFormer2025jbhi} & JBHI2025 & 62.50 & 66.31 & 58.62 & \multicolumn{1}{c|}{\underline{72.41}}  & \textbf{88.90} & \textbf{96.10} & \textbf{88.60} & \textbf{97.60} \\
Ours & 2026 & \textbf{86.29} & \textbf{80.65} & \textbf{91.94} & \multicolumn{1}{c|}{\textbf{87.71}}  & \underline{87.90} & \underline{92.16} & \underline{88.10} & \underline{91.54}\\
\bottomrule  
\end{tabular}
\end{table*}

\subsubsection{SIGF}
The Sequential Fundus Images for Glaucoma Forecast (SIGF) dataset \cite{DeepGF2020miccai} is the first publicly available longitudinal fundus images database for glaucoma forecasting. It contains 405 image sequences from different eyes, each with at least six time points collected between 1986 and 2018, totaling 3,671 images. All fundus images are annotated with binary glaucoma labels. Among the sequences, 37 progress from negative to positive glaucoma, while 368 remain negative throughout follow-up.

Previous studies followed the sampling strategy in \cite{DeepGF2020miccai} by generating overlapping clips, where each clip consists of five sequential input images and a sixth image serving as the prediction target. These clips are extracted from longer trajectories to augment the dataset while preserving temporal order. However, predicting glaucoma using input images that are already labeled positive provides limited clinical value. Therefore, we adopt a revised sampling strategy as show in Fig. \ref{fig4-4}. From each original sequence, we construct progression clips in which all input images are non-glaucoma, while the prediction target may be either glaucoma-positive or negative. To simulate realistic clinical follow-up intervals, we require that the last two time points be no more than three years apart. Constraining the interval encourages the model to focus on early structural cues within a clinically actionable time frame while reducing variability from irregular long-term follow-ups, which better reflects the clinical objective of forecasting disease before it becomes detectable. If the final image is labeled glaucoma, the sequence is designated as “progression”; if not, as “non-progression”. The data distribution of the progression sequences with different lengths is summarized in Table \ref{tab1}.


\subsubsection{GRAPE}
The Glaucoma Real-world Appraisal Progression Ensemble (GRAPE) dataset \cite{GRAPE} is a multi-modal longitudinal collection of visual field (VF) and fundus images for glaucoma. It includes 1,115 longitudinal records from 263 eyes of 144 patients collected between 2015 and 2022. The dataset contains VFs, fundus images, OCT measurements, and clinical information, with annotated optic disc segmentation and VF progression. The progression was defined using standard Octopus VF criteria based on point-wise linear regression (PLR2, PLR3) and mean deviation (MD) slope, confirmed by three glaucoma specialists. For our experiments, the first two fundus images from each eye were used as a sequence to predict progression, and sequences containing only one image are discard. The data distribution of sequences under three criteria is shown in Table \ref{tab2}.

\subsection{Evaluation Metrics}
We used the commonly used evaluation metrics including Accuracy (ACC), Sensitivity (SEN), Specificity (SPE), and Area Under the Receiver Operating Characteristic Curve (AUC) to evaluate the proposed method and the comparison methods. 
They can be expressed as follows:
\begin{equation}
    ACC = \frac{TP + TN} {TP + FN + TN + FP }
\end{equation}
\begin{equation}
     SEN= \frac{TP} {TP + FN} ,\space SPE= \frac{TN} {TN + FP}
\end{equation}
where TP, TN, FP, and FN represent the number of samples detected as true positive, true negative, false positive, and false negative.

In addition, we further used AUC to evaluate the performance, which is defined as the area under the Receiver Operating Characteristic (ROC) curve, which plots the true positive rate (SEN) against the false positive rate (1–SPE) at varying decision thresholds. AUC measures the overall discriminative ability of the model: an AUC of 0.5 corresponds to random guessing, while an AUC of 1.0 indicates perfect classification.

\subsection{Implementation Details}
Our method was implemented using the PyTorch framework, and all the experiments were conducted on NVIDIA RTX A6000 GPUs. During training, the performance was monitored on the validation set after each epoch. We trained our model for over 800 epochs and we observed that the validation loss plateaued after approximately 800 epochs. The batch size was fixed at 4, and optimization was performed using the AdamW optimizer with an initial learning rate of $3\times{10}^{-8}$, following a cosine annealing with warm restarts learning rate schedule. The RETFound{\_}mae{\_}natureCFP pre-trained weights were adopted to initialize the RETFound backbone.

\begin{table*}[]
\setlength{\tabcolsep}{3.5mm}
\centering
\caption{{Comparison of the proposed method with other methods on GRAPE dataset.(unit:\%)}}\label{tab4}
\begin{tabular}{clcclclcc}
\toprule
                               & \multicolumn{2}{c}{PLR2}            
                               & \multicolumn{2}{c}{PLR3}                   & \multicolumn{2}{c}{MD}          
                               & \multicolumn{2}{c}{Average} \\ 
    \cmidrule(r){2-3}   \cmidrule(r){4-5}   \cmidrule(r){6-7}   \cmidrule(r){8-9} 
\multicolumn{1}{c}{Method} & ACC & \multicolumn{1}{l|}{AUC}   & \multicolumn{1}{l}{ACC} & \multicolumn{1}{l|}{AUC}  & \multicolumn{1}{l}{ACC} & \multicolumn{1}{l|}{AUC} & \multicolumn{1}{l}{ACC} & AUC   \\  \midrule
\multicolumn{1}{c|}{GRAPE*}  & 75.00 & \multicolumn{1}{c|}{71.00}
& 91.00 & \multicolumn{1}{l|}{80.00} & 81.00 & \multicolumn{1}{l|}{73.00} 
& 82.33 & 74.67\\

\multicolumn{1}{c|}{GRAPE}     & 78.05 & \multicolumn{1}{c|}{66.45} 
& 82.18 & \multicolumn{1}{l|}{76.77} & 81.69  & \multicolumn{1}{l|}{\underline{79.07}}  & 80.64 & 74.10\\
\multicolumn{1}{c|}{Uniformer} & 78.10 & \multicolumn{1}{c|}{71.00} 
& 80.56 & \multicolumn{1}{l|}{87.40} & 83.27 & \multicolumn{1}{l|}{77.49}
& 80.65 & 78.55\\
\multicolumn{1}{c|}{GLIM-Net}  & \underline{80.09} & \multicolumn{1}{c|}{\underline{72.37}} 
& \underline{87.71} & \multicolumn{1}{l|}{\textbf{89.90}} & \underline{83.67} & \multicolumn{1}{l|}{76.39} & \underline{83.82} & \underline{79.55} \\
\multicolumn{1}{c|}{Ours}  & \textbf{81.18} & \multicolumn{1}{c|}{\textbf{79.14}} & \textbf{92.50}  & \multicolumn{1}{l|}{\underline{88.03}} & \textbf{88.77}  & \multicolumn{1}{l|}{\textbf{79.58}} & \textbf{87.46} & \textbf{82.25} \\ 
\bottomrule
\end{tabular}
\end{table*}

\section{Results}

\subsection{Comparison to the Sate-of-the-Art Methods}
\subsubsection{SIGF}
We compared with methods related to glaucoma forecasting and time-related sequence modeling, including ConvLSTM \cite{convlstm2015}, DeepGF \cite{DeepGF2020miccai}, Uniformer \cite{Uniformer2023}, GLIM-Net \cite{Glim2023tmi} and MST-Former \cite{MSTFormer2025jbhi}. ConvLSTM and DeepGF are CNN–LSTM-based methods that combine CNN-based feature extraction with LSTM-based temporal modeling. Uniformer, GLIM-Net and MST-Former are Transformer-based models. Uniformer integrates convolution and self-attention for spatiotemporal representation learning, which has achieved the leading performance on a variety of image and video classification tasks. GLIM-Net is a time-aware Transformer that incorporates time-positional encoding and time-sensitive attention to handle irregularly sampled fundus sequences. MST-Former is a multi-scale spatio-temporal Transformer that introduces space–time positional encoding and previously achieved state-of-the-art performance in glaucoma forecasting tasks. 
To ensure a fair comparison, all the comparison methods were trained and evaluated under the same experimental settings, using both the newly constructed progression sequences and the original augmented sequences.
The experimental results are shown in Table \ref{tab3}, with the best results shown in bold and the second best results underlined. $T$ stands for the number of time points used to forecast the next image.

For progression sequences ($T=2$), our method achieved the best overall performance with an ACC of 86.29\%, SEN of 80.65\%, SPE of 91.94\%, and AUC of 87.71\%, surpassing MST-Former by 23.79\% in ACC and 15.3\% in AUC. In this setting, all input images are non-glaucoma, making prediction particularly challenging due to the absence of explicit pathological cues. 
Existing CNN–LSTM and Transformer-based methods encode images as global representations and rely on temporal aggregation. Without visible disease patterns, these features are often dominated by appearance variability, e.g., illumination and imaging conditions, leading to unstable predictions and reliance on the most recent image. 
In contrast, DiffSight-Former explicitly models inter-visit structural differences in anatomically relevant regions, enabling the extraction of subtle yet consistent progression signals from normal-appearing inputs. This explains the notable gains in AUC and specificity, reflecting more reliable early-stage discrimination.

For augmented sequences ($T=5$), which include glaucoma-positive images, our method achieved an ACC of 87.90\% with competitive SEN and SPE at 92.16\% and 88.10\%, remaining close to the state-of-the-art. 
Although additional disease-related information is available, the performance gain of DiffSight-Former is more moderate than that of other methods. This is because many existing approaches benefit disproportionately from strong diagnostic cues in the most recent positive image, effectively shifting toward late-stage recognition rather than progression modeling. 
By contrast, DiffSight-Former emphasizes inter-visit structural change over static appearance, meaning that much of the discriminative information is already captured in the early forecasting setting. The inclusion of positive images therefore provides complementary but less dominant cues, resulting in more stable performance across settings.


\subsubsection{GRAPE}
We further evaluated the proposed method on the GRAPE dataset to assess the generalizability of the proposed method. Based on the results on SIGF dataset, we select several methods standing for Transformer-based structure for comparison. Table \ref{tab4} presents the comparison of our method against the existing models on the GRAPE dataset, with progression defined by Octopus VF criteria (PLR2, PLR3, and MD slope) and their average scores. The first row refers to the result of the validation baseline for VF progression prediction in their original paper, which only used the first image of each eye. For the rest rows, all the methods used the first and second images from each eye for prediction.

Under the PLR2 criterion, our method achieved the best performance with an ACC of 81.18\% and an AUC of 79.14\%, outperforming both baseline and competing methods. Under the stricter PLR3 definition, our approach attained the highest accuracy of 92.50\%. For the MD slope criterion, DiffSight-Former again achieved the best results, with an ACC of 88.77\% and an AUC of 79.58\%.

The consistent gains across these VF-based criteria indicate that the proposed model captures progression signals that are relevant to functional deterioration. Since visual field loss reflects accumulated functional damage that is often preceded by subtle structural changes in the optic nerve head and vasculature, modeling inter-visit structural variation provides a more reliable indicator of progression. By explicitly encoding these changes in anatomically meaningful regions, DiffSight-Former aligns more closely with the structure–function relationship in glaucoma, leading to stable performance across heterogeneous progression definitions.



\subsection{Ablation Study}
\subsubsection{Overall Structure}
To evaluate the contribution of each component in our proposed framework, we conducted ablation studies on SIGF dataset on three key modules: the Time-variant Feature Extraction (TFE) module, the Multi-Structure Difference Modulation (MDM) module, and the Time-Aware Transformer (TAT) module.

As presented in Table \ref{tab5}, the TFE backbone alone shows limited performance, indicating that feature extraction alone is insufficient for progression prediction, as it does not explicitly encode inter-visit change. 
Incorporating TAT improves performance to 83.87\% ACC and 79.81\% AUC, suggesting that modeling irregular temporal intervals enables the network to better weight observations across visits, particularly when follow-up spacing varies. 
Introducing MDM further boosts performance to SEN 92.76\% and AUC 86.76\%), highlighting that explicitly capturing structural differences in the optic disc/cup and vasculature makes subtle progression signals more detectable. 
The combination of all modules yields the best results, demonstrating that jointly modeling temporal dynamics and anatomically localized changes is essential for accurate glaucoma progression prediction.

\begin{table}
\setlength{\tabcolsep}{2.2mm}
\centering
\caption{Ablation study results on the overall structure.(unit:\%)}\label{tab5}
\begin{tabular}{cccccccc}
\toprule
      & TFE         & TAT       & MDM       & ACC   & SEN    & SPE   & AUC    \\ \midrule
1     & \checkmark  &           &           & 71.77 & 72.58  & 70.97 & 64.67   \\
2     & \checkmark  &\checkmark &           & 83.87 & 72.58  & 90.16 & 79.81  \\ 
3     & \checkmark  &           &\checkmark & 85.58 & \textbf{92.76}  & 78.39  &  86.76  \\ 

4     & \checkmark  &\checkmark &\checkmark & \textbf{86.29}  & 80.65 & \textbf{91.94} & \textbf{87.71}  \\ 
\bottomrule
\end{tabular}
\end{table}

\begin{figure}
\centering
\includegraphics[width=8cm]{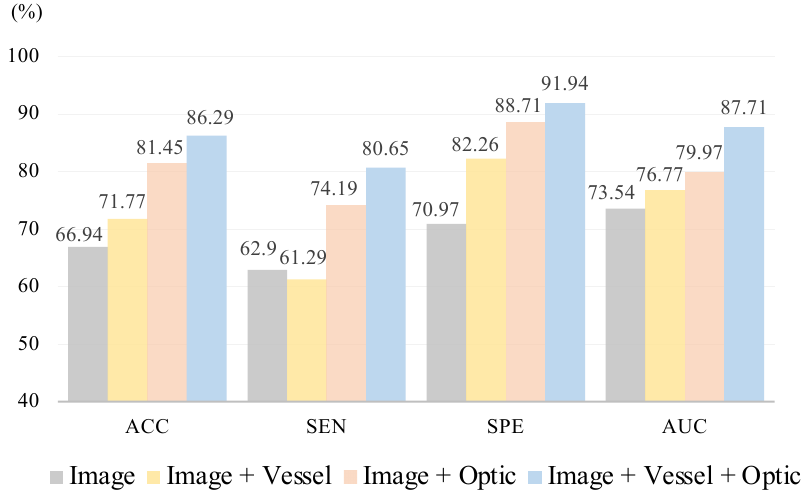}
\caption{Ablation study results on the Multi-structure Difference Modulation. Different structural maps is used to compute difference features. (unit:\%)} 
\label{fig4-1}
\end{figure}

\begin{figure*}
\centering
\includegraphics[width=17cm]{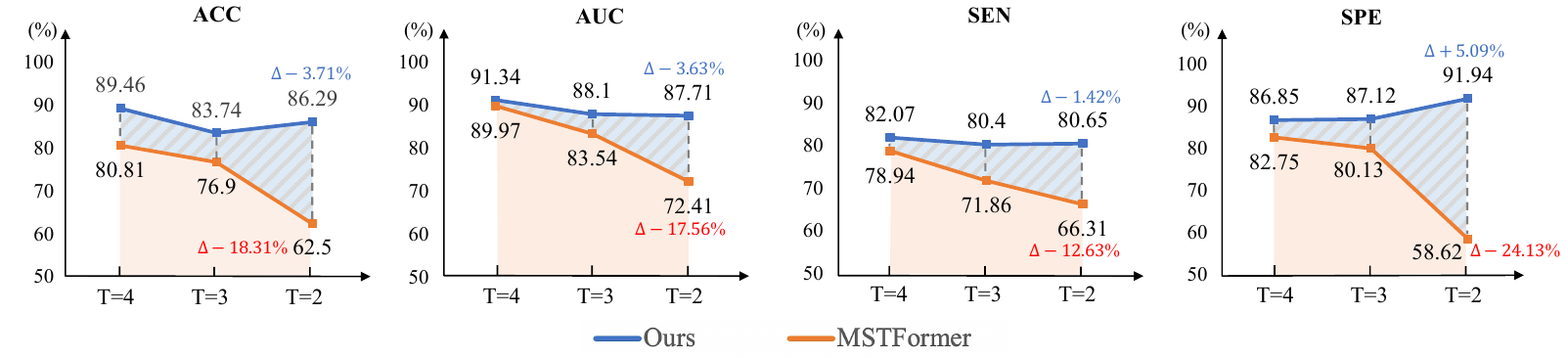}
\caption{The results comparison on different sequence length (T = 4, 3, 2) between our method and the MST-Former.}
\label{fig4-2}
\end{figure*}

\subsubsection{Multi-structure Difference Modulation}
To further assess the effectiveness of the proposed MDM module, we performed ablation experiments by incrementally incorporating different structural maps into the difference features computation. 
As shown in Fig. \ref{fig4-1}, using only the original image yields limited performance, indicating that appearance features alone are insufficient to capture progression signals. 
Incorporating either the vessel map or the optic disc/cup map leads to clear improvements, suggesting that progression-related changes are more prominently reflected in anatomically structured regions rather than the global fundus appearance. In particular, vessel maps capture vascular attenuation and distortion, while disc/cup maps emphasize morphological deformation, both of which are known indicators of glaucoma progression. 
When all three components are combined, the model achieves the best performance, demonstrating that these structural cues provide complementary information. This indicates that explicitly modeling inter-visit differences in multiple anatomical structures enables more reliable detection of subtle progression patterns that may not be observable from raw images alone.

\subsection{Sequence Length Analysis}
Fig. \ref{fig4-2} compares the performance of DiffSight-Former and MST-Former across different progression sequence lengths. MST-Former is selected as a representative state-of-the-art method due to its strong performance and comparable Transformer-based architecture.
When four images were used ($T=4$), our model achieved the highest AUC of 91.34\%, outperforming MST-Former, which achieved an AUC of 89.97\%. At three images ($T = 3$), both methods showed performance degradation; however, our method maintained a clear advantage with an AUC of 88.10\%, compared to 83.54\% for MST-Former. Even under the most challenging setting of only two available images ($T = 2$), DiffSight-Former achieved strong performance, obtaining an AUC of 87.71\% with balanced sensitivity of 80.65\% and specificity of 91.94\%. While MST-Former dropped markedly to an AUC of 72.41\%, a SEN of 66.31\% and a SPE of 58.62\%.


The superior performance of DiffSight-Former under reduced sequence lengths stems from how progression cues are extracted and aggregated. MST-Former primarily encodes each image as a global representation and infers progression through temporal aggregation. When only a few images are available, this approach becomes unreliable because global appearance changes are dominated by imaging variability rather than disease evolution. In contrast, DiffSight-Former is designed to directly isolate inter-visit anatomical change. The MDM module explicitly measures structural differences in disc/cup and vascular regions, enabling progression signals to be extracted even from a single interval, while the TFE module stabilizes feature representations across visits by suppressing non-pathological variations. As a result, the model depends less on long temporal context and more on localized, biologically meaningful change, which explains both its robustness and the observed improvement in specificity under short-sequence settings.



\begin{figure*}
\centering
\includegraphics[width=13cm]{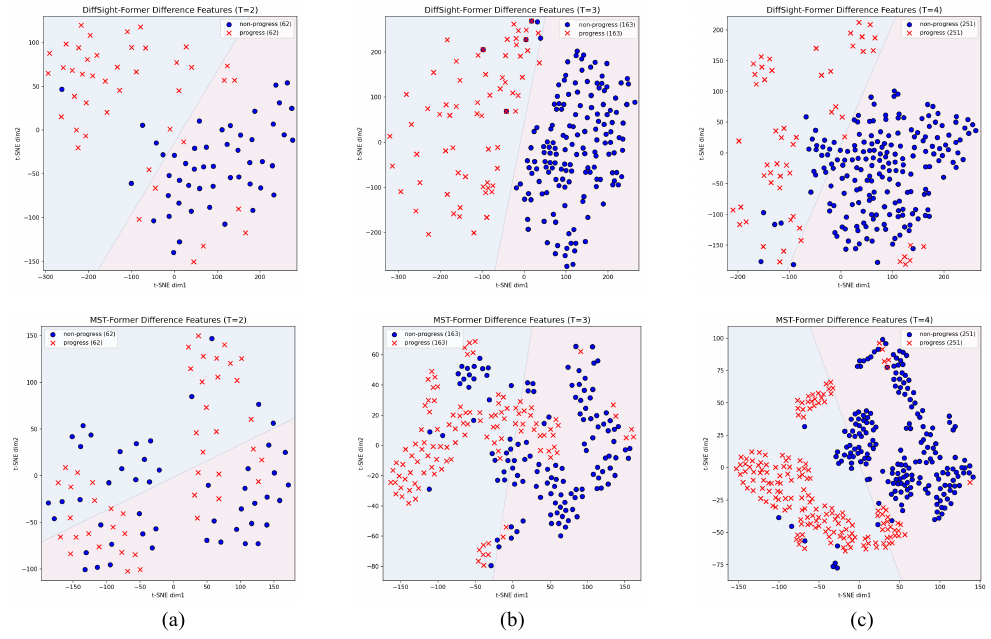}
\caption{The t-SNE distribution analysis of the difference features on test set under different temporal configurations produced by our method and MST-Former. In (a),(b),(c), the sequences used are T = 2,3,4 respectively.}
\label{fig4-3}
\end{figure*}

\subsection{Effectiveness of Difference Modeling}
To further evaluate the effectiveness of our proposed difference modeling, we conducted a feature-level comparison with MST-Former, the second best performing method. To ensure a fair comparison, we extracted intermediate features from the official MST-Former implementation by taking the encoded feature sequence from the last encoder layer and computing temporal difference features using the same formulation as in DiffSight-Former. This ensures that both models are evaluated under an identical feature definition, removing architectural biases and enabling an equitable analysis.

We applied t-SNE to project the high-dimensional difference features into a 2-dimensional space for the test sets under three temporal configurations of 2, 3, and 4 input images. As shown in Fig. \ref{fig4-3}, (a), (b), and (c) visualize the projected difference features for T = 2, 3, and 4, with DiffSight-Former in the top row and MST-Former in the bottom row. The representations produced by DiffSight-Former exhibit markedly clearer separation between progression and non-progression cases, forming well-defined and largely linearly separable clusters with a consistent separating direction across input lengths. As expected, both methods show slightly improved class separability when more time points are available, reflecting the additional temporal context. 


The observations suggest that the improved separability arises from how progression cues are encoded. While both methods utilize temporal information, MST-Former primarily aggregates global appearance features, which are sensitive to imaging variability and may not consistently reflect disease-related change. As a result, its difference features capture mixed signals dominated by appearance similarity rather than true progression patterns. In contrast, DiffSight-Former explicitly encodes inter-visit structural differences in anatomically relevant regions, allowing the resulting feature space to be organized according to progression-related change. This leads to more stable cluster geometry and a consistent separating direction across different input lengths. Such representations are more aligned with the underlying progression process in glaucoma, where subtle changes in the optic disc/cup and vasculature accumulate over time. 

These findings provide further evidence that modeling structured anatomical change, rather than relying on implicit temporal aggregation, is critical for learning meaningful progression representations, consistent with the performance gains observed in both ablation studies and main experiments.

\section{Discussion} 
This study demonstrates that explicitly modeling inter-visit structural differences provides a more effective strategy for glaucoma progression prediction than relying on global appearance representations. Across both datasets and experimental settings, DiffSight-Former consistently shows improved performance, particularly under early-stage conditions where input images are non-glaucoma. This suggests that progression signals are more reliably captured through localized anatomical changes, such as optic disc/cup deformation and vascular variation, rather than global image features. By decoupling structural change from appearance variability, the proposed framework reduces sensitivity to imaging noise and mitigates bias toward the most recent visit, leading to more stable and clinically meaningful predictions.

From a clinical perspective, the ability to predict progression from normal-appearing fundus images is particularly valuable for early intervention and long-term disease management. Unlike conventional approaches that depend on observable pathological signs, DiffSight-Former focuses on subtle progression-related changes that may precede clinical diagnosis. Furthermore, its consistent performance on the GRAPE dataset indicates that the learned structural representations generalize beyond image-level prediction to functional outcomes defined by visual field criteria. This alignment with the underlying structure–function relationship in glaucoma highlights the potential of the proposed framework for real-world clinical applications.

Despite these promising results, several limitations remain. The current study focuses on fundus image sequences, and future work will explore extending the framework to other longitudinal imaging modalities, such as OCT, as well as to other chronic diseases. In addition, validation on larger and more diverse clinical cohorts is needed to further assess robustness and facilitate translation into clinical practice.

\section{Conclusion}
We proposed DiffSight-Former, a framework for glaucoma progression prediction from longitudinal fundus image sequences. By explicitly modeling inter-visit structural differences and incorporating temporal dynamics, the proposed method captures subtle progression cues that are often missed by approaches relying on global appearance representations. 
Experiments on the SIGF and GRAPE datasets demonstrate consistent performance improvements over CNN–LSTM and Transformer-based baselines, with particularly strong performance under short-sequence settings where input images are non-glaucoma. These results highlight the importance of modeling progression as a difference-driven process rather than a static recognition task, which is critical for early-stage prediction in clinical practice. 
Future work will explore extending the framework to other longitudinal imaging modalities and validating its effectiveness on larger and more diverse multi-center datasets.

\section*{References}

\bibliographystyle{IEEEtran}
\bibliography{main.bib}

\end{document}